\documentclass[sigconf]{acmart}
\usepackage{multirow}
\usepackage{amssymb}
\usepackage{amsmath}
\usepackage{graphics}
\usepackage{graphicx}
\usepackage{array}
\usepackage{bm}
\usepackage{mathtools}

\AtBeginDocument{%
  \providecommand\BibTeX{{%
    \normalfont B\kern-0.5em{\scshape i\kern-0.25em b}\kern-0.8em\TeX}}}

\setcopyright{acmcopyright}
\copyrightyear{2019}
\acmYear{2019}
\acmDOI{}

\acmConference[recsysXfashion'19]{Workshop on Recommender Systems in Fashion, 13th ACM Conference on Recommender Systems}{September 20, 2019}{Copenhagen, Denmark}
\acmArticle{}
\acmPrice{}
\acmISBN{}

\begin{document}

\title{Attention-based Fusion for Outfit Recommendation}

\author{Katrien Laenen}
\email{katrien.laenen@kuleuven.be}
\orcid{}
\affiliation{%
  \institution{Department of Computer Science, KU Leuven,}
  \country{Belgium}
}

\author{Marie-Francine Moens}
\email{sien.moens@kuleuven.be}
\orcid{}
\affiliation{%
   \institution{Department of Computer Science, KU Leuven,}
   \country{Belgium}
}

%\renewcommand{\shortauthors}{Laenen et al.}

\begin{abstract}
This paper describes an attention-based fusion method for outfit recommendation which fuses the information in the product image and description to capture the most important, fine-grained product features into the item representation. We experiment with different kinds of attention mechanisms and demonstrate that the attention-based fusion improves item understanding.
We outperform state-of-the-art outfit recommendation results on three benchmark datasets.
\end{abstract}

\begin{CCSXML}
<ccs2012>
<concept>
<concept_id>10002951.10003260.10003282.10003550.10003555</concept_id>
<concept_desc>Information systems~Online shopping</concept_desc>
<concept_significance>500</concept_significance>
</concept>
<concept>
<concept_id>10010147.10010178.10010224.10010240.10010241</concept_id>
<concept_desc>Computing methodologies~Image representations</concept_desc>
<concept_significance>500</concept_significance>
</concept>
<concept>
<concept_id>10010147.10010178.10010224.10010245.10010255</concept_id>
<concept_desc>Computing methodologies~Matching</concept_desc>
<concept_significance>500</concept_significance>
</concept>
<concept>
<concept_id>10010147.10010178.10010179</concept_id>
<concept_desc>Computing methodologies~Natural language processing</concept_desc>
<concept_significance>100</concept_significance>
</concept>
</ccs2012>
\end{CCSXML}

\ccsdesc[500]{Information systems~Online shopping}
\ccsdesc[500]{Computing methodologies~Image representations}
\ccsdesc[500]{Computing methodologies~Matching}
\ccsdesc[100]{Computing methodologies~Natural language processing}

\keywords{outfit recommendation, item understanding, attention, attention-based fusion}

\begin{teaserfigure}
  \centering
  \includegraphics[width=0.5\textwidth]{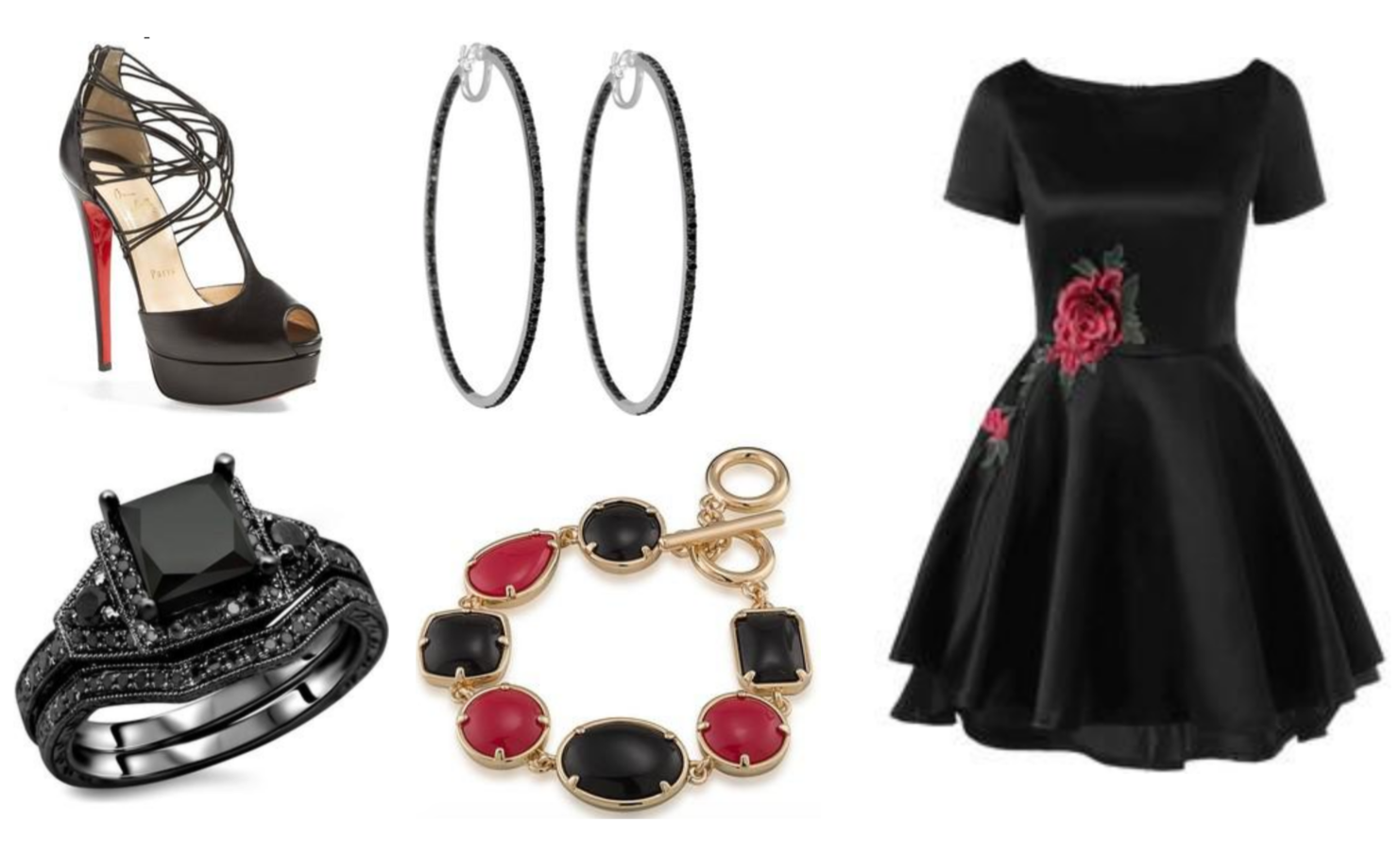}
  \caption{Example outfit in the Polyvore68K dataset. Fine details, such as the heels of the sandals, the flower applique on the dress and the red pendants of the bracelet, determine that these items match nicely. These details should therefore be captured in the item representations.}
  \label{fig:teaser}
\end{teaserfigure}

\maketitle

\section{Introduction}

With the explosive growth of e-commerce content on the Web, recommendation systems are essential to overcome consumer over-choice and to improve user experience. Often users shop online to buy a full outfit or to buy items matching other items in their closet. Webshops currently only offer limited support for these kinds of searches. Some webshops offer a \textit{people also bought} feature as suggestions for compatible clothing items. However, items that are bought together by others are not necessarily compatible with each other, nor do they necessarily correspond with the taste and style of the current user. Another feature some webshops provide is \textit{shop the look}. This enables to buy all clothing items worn together with the viewed item in an outfit which is usually put together by a fashion stylist. However, this scenario does not provide alternatives that might appeal more to the user.

In this work, we tackle the problem of outfit recommendation. The goal of this task is to compose a fashionable outfit either from scratch or starting from an incomplete set of items. Outfit recommendation has two main challenges. The first is \textit{item understanding}. Fine details in the garments can be important for making combinations. For example, the items in Figure \ref{fig:teaser} match nicely because of the red heels of the sandals, the red flowers on the dress and the red pendants of the bracelet. These fine-grained product details should be captured in the item representations. Moreover, usually there is also a short text description associated with the product image. These descriptions point out certain product features and contain information which is useful for making combinations as well. Hence, there is a need to effectively integrate the visual and textual item information into the item representations. The second challenge in outfit recommendation is \textit{item matching}. Item compatibility is a complex relation. For instance, assume items $A$ and $B$ are both compatible with item $C$. In that case items $A$ and $B$ can be, but are not necessarily, visually similar. Moreover, items $A$ and $B$ can be, but are not necessarily, also compatible with each other. Furthermore, different product features can play a role in determining compatibility depending on the types of items being matched, as illustrated in \cite{vasileva2018}.

This work will focus on \textit{item understanding}. Our outfit recommendation system operates on region-level and word-level representations to bring product features which are important to make item combinations to the forefront as needed. 
The contributions of our work are threefold.
Firstly, our approach works on a finer level of image regions and words. In contrast, previous approaches to outfit recommendation work on a more coarse level of full images and sentences.
Secondly, we explore different attention mechanisms and propose an attention-based fusion method which fuses the visual and textual information to capture the most relevant product features into the item representations. Attention mechanisms have not yet been explored in outfit recommendation systems to improve item understanding.
Thirdly, we improve state-of-the-art outfit recommendation results on three datasets.

The remainder of this paper is structured as follows. In Section \ref{sec:relatedwork} we review other works on outfit recommendation. Then, Section \ref{sec:methodology} describes our model architecture. Next, Section \ref{sec:experiments} contains our experimental setup. The results of the conducted experiments are analysed in Section \ref{sec:results}. Finally, Section \ref{sec:conclusion} provides our conclusions and directions for future work.

\section{Related Work}
\label{sec:relatedwork}

The task of outfit fashionability prediction requires to uncover which items go well together based on item style, color and shape.  This can be learned from visual data, language data or a combination of the two. Currently, two approaches are common to tackle outfit fashionability prediction. 
The first one is to infer a feature space where visually compatible clothing items are close together. \cite{veit2015} use a Siamese convolutional neural network (CNN) architecture to infer a compatibility space of clothing items. Instead of only one feature space, multiple feature spaces can also be learned to focus on certain compatibility relationships. \cite{rhe2016} propose to learn a compatibility space for different types of relatedness (e.g., color, texture, brand) and weight these spaces according to their relevance for a particular pair of items. \cite{vasileva2018} infer a compatibility space for each pair of item types (i.e., tops and bottoms, tops and handbags) and demonstrate that the embeddings specialize to features that dominate the compatibility relationship for that pair of types. Moreover, their approach also uses the textual descriptions of items to further improve the results. 
The second common approach to outfit fashionability prediction is to obtain outfit representations and to train a fashionability predictor on these outfit representations. In \cite{simoserra2015} a conditional random field scores the fashionability of a picture of a person's outfit based on a bag-of-words representation of the outfit and visual features of both the scenery and person. Their method also provides feedback on how to improve the fashionability score. In \cite{li2017} neural networks are used to acquire multimodal representations of items based on the item image, category and title, to pool these into one outfit representation and to score the outfit's fashionability.
Other approaches to outfit fashionability prediction also exist. In \cite{han2017} an outfit is treated as an ordered sequence and a bidirectional long short-term memory (LSTM) model is used to learn the compatibility relationships among the fashion items. In \cite{hsiao2018} the visual compatibility of clothing items is captured with a correlated topic model to automatically create capsule wardrobes. \cite{lin2019} build an end-to-end learning framework that improves item recommendation with co-supervision of item generation. Given an image of a top and a description of the requested bottom (or vice versa) their model composes outfits consisting of one top piece and one bottom piece.

None of the above approaches work with region-level and word-level representations, nor make use of an attention mechanism. In contrast, we infer which product features are most important for the outfit recommendation task through the use of an attention mechanism on regions and words.

\section{Methodology}
\label{sec:methodology}

Section \ref{ssec:baseline} describes our baseline model, which fuses the visual and textual information with standard common space fusion. Next, Section \ref{ssec:attention_fusion} elaborates our model architecture which fuses the visual and textual information through attention.

In all formulas, matrices are written with capital letters and vectors are bolded. We use letters $W$ and $\boldsymbol{b}$ to refer to respectively the weights and bias in linear and non-linear transformations.

\subsection{Baseline}
\label{ssec:baseline}

Our baseline model is the method of \cite{vasileva2018}.
The model receives two triplets as input: a triplet of image embeddings $(\boldsymbol{x}_{(u)}, \boldsymbol{x}^+_{(v)}, \boldsymbol{x}^-_{(v)})$ of dimension $d_i$ and a triplet of corresponding sentence embeddings $(\boldsymbol{t}_{(u)}, \boldsymbol{t}^+_{(v)}, \boldsymbol{t}^-_{(v)})$ of dimension $d_t$. How these image and sentence embeddings are obtained is detailed in Section \ref{ssec:training_details}. Embeddings $\boldsymbol{x}_{(u)}$ and $\boldsymbol{x}^+_{(v)}$ represent images of respectively type $u$ and type $v$ which are compatible. Compatible means that the images represented by $\boldsymbol{x}_{(u)}$ and $\boldsymbol{x}^+_{(v)}$ appear together in some outfit. Meanwhile $\boldsymbol{x}^-_{(v)}$ represents a randomly sampled image of the same type as $\boldsymbol{x}^+_{(v)}$ that has not been seen in an outfit with $\boldsymbol{x}_{(u)}$ and is therefore considered to be incompatible with $\boldsymbol{x}_{(u)}$.

The triplets are first projected to a common, semantic space $\mathcal{S}$ of dimension $d_g$. The purpose of the common space is to better capture the notions of image similarity, text similarity and image-text similarity. Therefore, three losses are defined on the common space.
A visual-semantic loss $\mathcal{L}_{vse}$ enforces that each image should be closer to its own description than to the descriptions of the other images in the triplet:
\begin{align}
    \mathcal{L}_{vse} &=
        \frac{
            \mathcal{L}_{vse,\boldsymbol{x}_{(u)}} + \mathcal{L}_{vse,\boldsymbol{x}^+_{(v)}} + \mathcal{L}_{vse,\boldsymbol{x}^-_{(v)}}}
            {3}
        \label{eq:vse}\\
    \mathcal{L}_{vse,\boldsymbol{x}_{(u)}} &=
        \frac{\begin{array}{@{}r@{}}
            \ell(W_i\boldsymbol{x}_{(u)}, W_s\boldsymbol{t}_{(u)}, W_s\boldsymbol{t}^+_{(v)}) \\ {}+ \ell(W_i\boldsymbol{x}_{(u)}, W_s\boldsymbol{t}_{(u)}, W_s\boldsymbol{t}^-_{(v)})
            \end{array}} 
            {2}
        \label{eq:vse_x}\\
        \text{with }
            &\ell(\boldsymbol{x}, \boldsymbol{y}, \boldsymbol{z}) = \max(0, f(\boldsymbol{x}, \boldsymbol{z}) - f(\boldsymbol{x}, \boldsymbol{y}) + m)\label{eq:tripletloss}\\
        \text{and  }
            &f(\boldsymbol{x}, \boldsymbol{y}) = \frac{\boldsymbol{x}^T\cdot\boldsymbol{y}}{||\boldsymbol{x}||\cdot||\boldsymbol{y}||}
\end{align}
with $W_i \in \mathbb{R}^{d_g \times d_i}$ and $W_s \in \mathbb{R}^{d_g \times d_t}$ projections to the common space, $\ell$ the standard triplet loss, $m$ the margin, and $f$ the cosine similarity. $\mathcal{L}_{vse,\boldsymbol{x}^+_{(v)}}$ and $\mathcal{L}_{vse,\boldsymbol{x}^-_{(v)}}$ are computed analogous to Eq. \ref{eq:vse_x}.
A visual similarity loss $\mathcal{L}_{vsim}$ enforces that an image of type $v$ should be closer to an image of the same type $v$ than to an image of another type $u$:
\begin{align}
    \mathcal{L}_{vsim} &=
        \frac{\begin{array}{@{}r@{}}
            \ell(W_i\boldsymbol{x}^+_{(v)}, W_i\boldsymbol{x}^-_{(v)}, W_i\boldsymbol{x}_{(u)}) \\ {}+ \ell(W_i\boldsymbol{x}^-_{(v)}, W_i\boldsymbol{x}^+_{(v)}, W_i\boldsymbol{x}_{(u)})
            \end{array}} 
            {2}\label{eq:vsimloss}
\end{align}
with $W_i \in \mathbb{R}^{d_g \times d_i}$ the image projection to the common space and $\ell$ the standard triplet loss of Eq. \ref{eq:tripletloss}. Finally, a textual similarity loss $\mathcal{L}_{tsim}$ is defined analogous to Eq. \ref{eq:vsimloss}.

Next, a type-specific compatibility space $\mathcal{C}_{(u, v)}$ of dimension $d_c$ is inferred for each pair of types $u$ and $v$. In $\mathcal{C}_{(u, v)}$ a compatibility loss $\mathcal{L}_{comp}$ enforces that compatible images are closer together than non-compatible images:
\begin{align}
    \mathcal{L}_{comp} &= \ell(W_c^{(u,v)}W_i\boldsymbol{x}_{(u)}, W_c^{(u,v)}W_i\boldsymbol{x}^+_{(v)}, W_c^{(u,v)}W_i\boldsymbol{x}^-_{(v)})\label{eq:vcomploss}
\end{align}
with $W_i \in \mathbb{R}^{d_g \times d_i}$ the image projection to the common space, $W_c^{(u,v)} \in \mathbb{R}^{d_c \times d_g}$ the projection associated with $\mathcal{C}_{(u, v)}$, and $\ell$ the standard triplet loss of Eq. \ref{eq:tripletloss}.

The final training loss is:
\begin{align}
    \mathcal{L} = \mathcal{L}_{comp} + \lambda_1 \mathcal{L}_{vsim} + \lambda_2 \mathcal{L}_{tsim} +\lambda_3 \mathcal{L}_{vse}
\end{align}
with $\lambda_1$, $\lambda_2$ and $\lambda_3$ scalar parameters.

\subsection{Attention-based Fusion for Outfit Recommendation}
\label{ssec:attention_fusion}

The downside of the baseline model is that the item representations are quite coarse and the interaction between the visual and textual modality is quite limited. Instead, we would like to highlight certain parts of an image or words in a description which correspond to important product features for making fashionable item combinations, and integrate this into a multimodal item representation. Therefore we propose an attention-based fusion model, which we obtain by making a few adjustments to the baseline model.

Firstly, the first input to the attention-based fusion model is a triplet of region-level image features $(\boldsymbol{x}_{\bm{1:N}(u)}, \boldsymbol{x}^+_{\bm{1:N}(v)}, \boldsymbol{x}^-_{\bm{1:N}(v)})$ of dimension $d_i$, where $N$ denotes the number of regions. Depending on the attention mechanism used, the other input is either a triplet of description-level features $(\boldsymbol{t}_{(u)}, \boldsymbol{t}^+_{(v)}, \boldsymbol{t}^-_{(v)})$ as before or a triplet of word-level features $(\boldsymbol{t}_{\bm{1:M}(u)}, \boldsymbol{t}^+_{\bm{1:M}(v)}, \boldsymbol{t}^-_{\bm{1:M}(v)})$ of dimension $d_t$, where $M$ denotes the number of words. Details on how these features are obtained can be found in Section \ref{ssec:training_details}.
Since $\mathcal{L}_{vsim}$ and $\mathcal{L}_{vse}$ are formulated at the level of full images, we obtain image-level representations by simply taking the average of the region-level representations, i.e., $\boldsymbol{x}_{(u)} = \frac{1}{N}\sum_{i=1}^{N}\boldsymbol{x}_{\bm{i}(u)}$. In the same way we obtain description-level representations from word-level representations for $\mathcal{L}_{tsim}$.

Secondly, we use an attention mechanism to fuse the visual and textual information and obtain a triplet $(\boldsymbol{m}_{(u)}, \boldsymbol{m}^+_{(v)}, \boldsymbol{m}^-_{(v)})$ of multimodal item representations. These multimodal item representations are more fine-grained and allow more complex interactions between the vision and language data. Finally, we project these multimodal item representations to the type-specific compatibility spaces.

How we identify important product features depends on the attention mechanism used. Section \ref{sssec:vdot_product_attention} describes visual dot product attention. Section \ref{sssec:stacked_vis_attention} describes stacked visual attention. Finally, Section \ref{sssec:mfb_coattention} discusses a co-attention mechanism. Furthermore, we also experimented with self-attention \cite{vaswani2017} on the image regions and words, and some other co-attention and multimodal attention mechanisms \cite{lu2016,seo2017,nam2017}, but these did not improve performance.

\subsubsection{Visual Dot Product Attention}
\label{sssec:vdot_product_attention}

Given region-level image features $X \in \mathbb{R}^{N \times d_g}$ and description-level features  $\boldsymbol{t} \in \mathbb{R}^{d_g}$, visual dot product attention produces attention weights based on the dot product of the representations of the description and each region:
\begin{align}
    a_i & =  \tanh(\boldsymbol{x_i}) \cdot \tanh(\boldsymbol{t})
\end{align}
with $\boldsymbol{x_i}$ the $i$'th row of $X$.
Next, the attention weights are normalized and used to compute the visual context vector:
\begin{equation}
    \boldsymbol{c} = \sum_{i=1}^{N} \alpha_{i} \boldsymbol{x_i} \text{, with } \alpha_{i} = \text{softmax}([a_1, a_2, ..., a_N])_i
\end{equation}
with $\boldsymbol{x_i}$ the $i$'th row of $X$. The visual context vector $\boldsymbol{c}$ is concatenated with description $\boldsymbol{t}$, i.e., $[\boldsymbol{c}; \boldsymbol{t}]$ with $[]$ the concatenation operator, to obtain a multimodal item representation of dimension $2d_g$.

\subsubsection{Stacked Visual Attention}
\label{sssec:stacked_vis_attention}

Given region-level image features $X \in \mathbb{R}^{N \times d_g}$ and description-level features  $\boldsymbol{t} \in \mathbb{R}^{d_g}$, stacked visual attention \cite{yang2016} produces a multimodal context vector in multiple attention hops, each extracting more fine-grained visual information.
In the $r$'th attention hop, the attention weights and context vector are calculated as:
\begin{align}
    \boldsymbol{a}^{(r)} &=  \boldsymbol{w}^{(r)}_p \tanh(W^{(r)}_v X^T \oplus (W^{(r)}_t \boldsymbol{q}^{{(r-1)}} + \boldsymbol{b}^{(r)}_s))\\
    \boldsymbol{c}^{(r)} &= \boldsymbol{\alpha}^{(r)}X \text{, with } \boldsymbol{\alpha}^{(r)} = \text{softmax}(\boldsymbol{a}^{(r)})
\end{align}
with $W^{(r)}_v, W^{(r)}_t \in \mathbb{R}^{h \times d_g}$ and $\boldsymbol{w}^{(r)}_p \in \mathbb{R}^{1 \times h}$ learnable weights, $\boldsymbol{b}^{(r)}_s \in \mathbb{R}^{h}$ the bias vector, $\boldsymbol{q}^{(r-1)}$ the query vector from the previous hop, and $\oplus$ the elementwise sum operator.
The query vector is initialized to $\boldsymbol{t}$. At the $r$'th hop, the query vector is updated as:
\begin{equation}
    \boldsymbol{q}^{(r)} = \boldsymbol{q}^{(r-1)} + \boldsymbol{c}^{(r)} 
\end{equation}
This process is repeated $R$ times, with $R$ the number of attention hops. Afterwards, the final query vector $\boldsymbol{q}^{(R)}$ is concatenated with description $\boldsymbol{t}$, i.e., $[\boldsymbol{q}^{(R)}; \boldsymbol{t}]$ with $[]$ the concatenation operator, to obtain a multimodal item representation of dimension $2d_g$.

\subsubsection{Co-attention}
\label{sssec:mfb_coattention}

The co-attention mechanism of \cite{yu2017} attends to both the representations of the image regions $X \in \mathbb{R}^{N \times d_g}$ and the representations of the description words $Y \in \mathbb{R}^{M \times d_g}$ as follows.

First, the description words are attended independent of the image regions. The assumption here is that the most relevant words of the description can be inferred independent of the image content, i.e., words referring to color, shape, style and brand can be considered relevant independent of whether they are displayed in the image or not. Given word-level features $Y$, the textual attention weights $\boldsymbol{a}^t$ and textual context vector $\boldsymbol{c}^t$ are obtained as:
\begin{align}
    \boldsymbol{a}^t &= \underset{in=d_g,out=1,k=1}{\text{Convolution1D}_{t,2}}(\text{ReLU}(\underset{in=d_g,out=d_g,k=1}{\text{Convolution1D}_{t,1}}(Y)))\\
    \boldsymbol{c}^t &= \boldsymbol{\alpha}^t Y \text{, with } \boldsymbol{\alpha}^t = \text{softmax}(\boldsymbol{a}^t)
\end{align}
where \textit{Convolution1D} refers to the 1D-convolution operation with \textit{in} input channels, \textit{out} output channels and kernel size \textit{k}.

Next, the image regions are attended in $R$ attention hops.
In the $r$'th attention hop, the textual context vector $\boldsymbol{c}^t$ is merged with each of the region-level image features in $X$ using multimodal factorized bilinear pooling (MFB). MFB consists of an \textit{expand stage} where the unimodal representations are projected to a higher dimensional space of dimension $p2d_g$ (with $p$ a hyperparameter) and then merged with elementwise multiplication followed by a \textit{squeeze stage} where the merged feature is transformed back to a lower dimension $2d_g$. For a detailed explanation of MFB the reader is referred to \cite{yu2017}. The MFB operation results in a multimodal feature matrix $M \in \mathbb{R}^{N \times 2d_g}$. Then, the visual attention weights $\boldsymbol{a}^{v,(r)}$ and context vector $\boldsymbol{c}^{v,(r)}$ are calculated based on this merged multimodal feature matrix $M$:
\begin{align}
    \boldsymbol{a}^{v,(r)} &= \underset{in=d_g,out=1,k=1}{\text{Convolution1D}^{(r)}_{v,2}}(\text{ReLU}(\underset{in=2d_g,out=d_g,k=1}{\text{Convolution1D}^{(r)}_{v,1}}(M)))\\
    \boldsymbol{c}^{v,(r)} &= \boldsymbol{\alpha}^{v,(r)} M \text{, with } \boldsymbol{\alpha}^{v,(r)} = \text{softmax}(\boldsymbol{a}^{v,(r)})
\end{align}
The visual context vectors of all hops are concatenated and transformed to obtain the final visual context vector $\boldsymbol{c}^{v}$:
\begin{equation}
    \boldsymbol{c}^{v} =  W_f [\boldsymbol{c}^{v,(1)}; \boldsymbol{c}^{v,(2)}; ...; \boldsymbol{c}^{v,(R)}]
\end{equation}
with $W_f \in \mathbb{R}^{R2d_g \times 2d_g}$ and $[]$ the concatenation operator.
Finally, the final visual context vector $\boldsymbol{c}^{v}$ is merged with the textual context vector $\boldsymbol{c}^{t}$ using MFB to acquire a multimodal item representation of dimension $2d_g$.

\section{Experimental Setup}
\label{sec:experiments}

\subsection{Experiments and Evaluation}
\label{ssec:experiments_and_evaluation}

All models are evaluated on two tasks. In the fashion compatibility (FC) task, a candidate outfit is scored based on how compatible its items are with each other. More precisely, the outfit compatibility score is computed as the average compatibility score across all item pairs in the outfit. Since the compatibility of two items is measured with cosine similarity, the outfit compatibility score will lie in the interval $[-1, 1]$. The performance of the FC task is evaluated using the area under a ROC curve (AUC). In the fill-in-the-blank (FITB) task the goal is to select from a set of four candidate items the item which is the most compatible with the remainder of the outfit. More precisely, the most compatible candidate item is the one which has the highest total compatibility score with the items in the remainder of the outfit. Performance for this task is evaluated with accuracy.

FC questions and FITB questions that consist of images without a description are discarded to keep evaluation fair for all models.
Also note that if a pair of items have a type combination that was never seen during training, the model has not learned a type-specific compatibility space for that pair. Such pairs are ignored during evaluation. Hence, we also use the training set to determine which pairs of types do not effect outfit fashionability.

\subsection{Datasets}

We evaluate all models on three different datasets: Polyvore68K-ND, Polyvore68K-D and Polyvore21K.

\subsubsection{Polyvore68K}

The Polyvore68K dataset\footnote{https://github.com/mvasil/fashion-compatibility} \cite{vasileva2018} originates from Polyvore. Two different train-test splits are defined for the dataset. Polyvore68K-ND contains 53,306 outfits for training, 10,000 for testing, and 5,000 for validation. It consists of 365,054 items, some of which occur both in the training and test set. However, no outfit appearing in one of the three sets is seen in the other two.
The other split, Polyvore68K-D, contains 32,140 outfits, of which 16,995 are used for training, 15,145 for testing and 3,000 for validation. It has 175,485 items in total, where no item seen during training appears in the validation or test set. Both splits have their own FC questions and FITB questions.

Each item in the dataset is represented by a product image and a short description. Items have one of 11 coarse types (see Table \ref{tab:types} in Appendix A).

\subsubsection{Polyvore21K}
\label{sssec:Polyvore21K}

Another dataset collected from Polyvore is the Polyvore21K dataset\footnote{https://github.com/xthan/polyvore-dataset} \cite{han2017}. It contains items of 380 different item types, however not all are fashion related, e.g., furniture, toys, skincare, food and drinks, etc. We delete all items with types unrelated to clothing, clothing accessories, shoes and bags.
The remaining 180 types are all fashion related, but some of them are very-fine grained. We make the item types more coarse to avoid an abundance of type-specific compatibility spaces, i.e. more than 5,000, which is unfeasible. The remaining 37 types can be found in Table \ref{tab:types} in Appendix A. Eventually, this leaves 16,919 outfits for training, 1,305 for validation and 2,701 for testing. There are no overlapping items between the three sets. Each item has an associated image and description.

During evaluation we use the FC questions and FITB questions supplied by \cite{vasileva2018} for the  Polyvore21K dataset, after removal of fashion unrelated items.

\subsection{Comparison with Other Works}

This work uses a slightly different setup than the work of \cite{vasileva2018} and therefore our results are not exactly comparable with theirs. Firstly, we do not evaluate our models on the same set of FC and FITB questions. This is because we discard questions consisting of images without a description as explained in Section \ref{ssec:experiments_and_evaluation}. Secondly, the item types used for the Polyvore21K dataset are different. It is unclear from \cite{vasileva2018} how they obtain and use the item types of the Polyvore21K dataset, as these have only been made public recently. In this work, we used the publicly available item types after cleaning as detailed in Section \ref{sssec:Polyvore21K}.

\subsection{Training Details}
\label{ssec:training_details}

\begin{figure*}
  \centering
  \includegraphics[width=\textwidth]{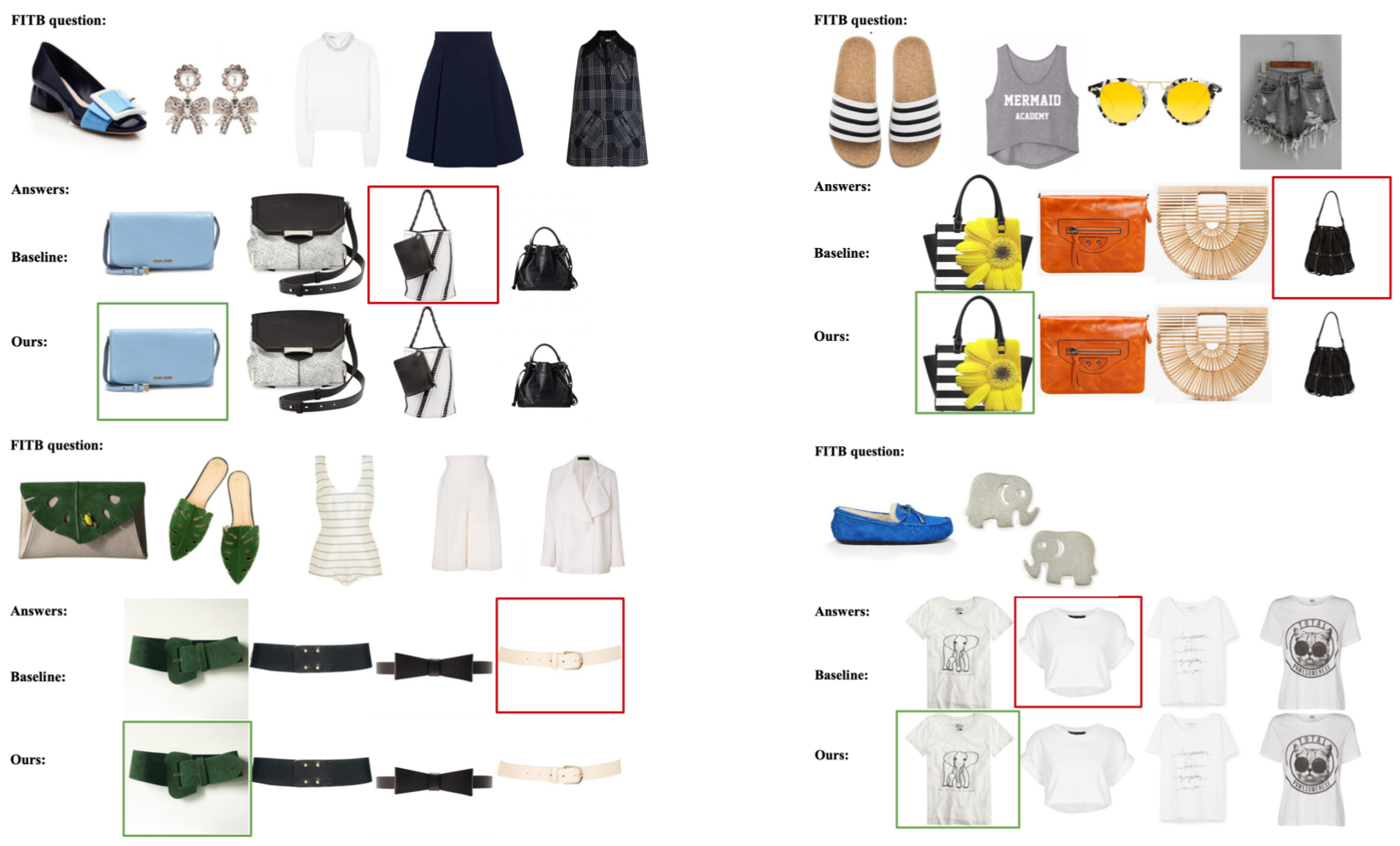}
  \caption{Examples of fill-in-the-blank questions on the Polyvore68K-ND dataset and answers generated by the baseline model and our attention-based fusion model based on stacked visual attention.}
  \label{fig:results}
\end{figure*}

All images are represented with the ResNet18 architecture \cite{he2016} pretrained on ImageNet. More precisely, as in \cite{vasileva2018} we take the output of the $7\times7\times256$ \textit{res4b\_relu} layer. For the models operating on image regions this results in 49 regions for every image, each with a dimension $d_i$ of 256. For the models working with full images, we use an additional average pooling layer to obtain one image-level representation, also with a dimension $d_i$ equal to 256.
The text descriptions are represented with a bidirectional LSTM of which the forward and backward hidden state at timestep $M$ are concatenated, with $M$ the number of words in the descriptions. For models operating on the level of words instead of full descriptions, we concatenate the forward and backward hidden state of the bidirectional LSTM at each timestep $j$ to obtain the representation for the $j$'th word. The parameters of the ResNet18 architecture and the bidirectional LSTM are finetuned on our dataset during training. Dimensions $d_t$, $d_g$, $d_c$ and $h$ are equal to 512. Hyperparameters are set based on the validation set. For the attention mechanisms, the number of attention hops $R$ is set to 2 and hyperparameter $p$ for MFB is set to 2.

All models are trained for $10$ epochs using the ADAM optimizer with a learning rate of 5e-5 and a batch size of 128. In the loss functions, factors $\lambda_1$ and $\lambda_2$ are 5e-5, $\lambda_3$ is set to 5e-3 and margin $m$ is $0.2$.
All models are trained for 5 runs. We do this to counteract the effect of the negative sampling which is done at random during training. To compute performance, we take the average performance on the FC task and FITB task across these 5 runs. In qualitative results, we use a voting procedure to determine the final answer on FC and FITB questions.

\section{Results}
\label{sec:results}

\begin{table}[t]
    \centering
    \begin{center}
    \resizebox{0.48\textwidth}{!}{%
    \begin{tabular}{| l l | c c | c c | c c |}
    \hline
     & & \multicolumn{2}{c|}{\textbf{Polyvore68K-ND}} & \multicolumn{2}{c|}{\textbf{Polyvore68K-D}} & \multicolumn{2}{c|}{\textbf{Polyvore21K}} \\
     & & FC & FITB & FC & FITB & FC & FITB \\
     \hline
     \multicolumn{2}{|l|}{\textit{Common space fusion}} &  &  &  &  &  & \\
     & baseline \cite{vasileva2018} & 85.62 & 56.55 & 85.07 & 56.91 & 86.28 & 58.35 \\ 
     \hline
     \multicolumn{2}{|l|}{\textit{Attention-based fusion}} &  &  &  &  &  & \\
     & visual dot product attention & 89.43 & 61.55 & 86.85 & 60.12 & 88.59 & \textbf{63.11} \\
     & stacked visual attention & \textbf{89.68} & \textbf{61.92} & \textbf{87.25} & \textbf{60.48} & \textbf{88.89} & 62.52 \\
     & co-attention & 89.58 & 61.20 & 86.25 & 59.00 & 85.04 & 58.20 \\
    \hline
    \end{tabular}}
    \end{center}
    \caption{Results on the fashion compatibility and fill-in-the-blank tasks for the Polyvore68K dataset versions and the Polyvore21K dataset.}
    \label{tab:results}
\end{table}

Table \ref{tab:results} shows the results of the discussed models on the Polyvore68K dataset versions and the Polyvore21K dataset. We outperform standard common space fusion on all three datasets for both the FC and FITB tasks.
On the Polyvore68K dataset versions the best results for both tasks are achieved with the fusion method based on stacked visual attention.
For the Polyvore21K dataset the best results for the FC task are obtained with the fusion method based on stacked visual attention and for the FITB task with the fusion method based on visual dot product attention.
Generally, we observe that a basic attention mechanism such as visual dot product attention obtains comparable results with more complex attention mechanisms such as stacked visual attention or co-attention.

When focusing on the separate tasks, our attention-based fusion models seem better at distinguishing randomly generated outfits from human-generated outfits than the standard common space fusion models. Especially on the Polyvore68K-ND dataset this observation is apparent.
Furthermore, our attended multimodal item representations enable the generation of more fashionable outfits as can be seen from the results on the FITB task.
Figure \ref{fig:results} shows some FITB questions and answers generated by the standard common space fusion model and our fusion model based on stacked visual attention for the Polyvore68K-ND dataset. For each of these FITB questions, the ground truth item needs to be selected because of some small details in other items of the outfit which are picked up by our model but not by the baseline. 
More precisely, for the first example the light blue handbag matches especially well with the light blue clasp of the pump. In the second example, the striped pattern of the handbag returns in the slippers and the yellow of the flower on the handbag returns in the sunglasses. In the third example, the green belt matches well with the green accents in the handbag and mules. In the last example, the T-shirt of an elephant looks nice in combination with the elephant-shaped earrings.

Hence, both quantitative and qualitative results demonstrate that highlighting certain product features in the item representations for making outfit combinations is meaningful and can be achieved with attention.

\section{Conclusion}
\label{sec:conclusion}

In this work we showed that attention-based fusion integrates visual and textual information in a more meaningful way than standard common space fusion.
Attention on region-level image features and word-level text features allows to bring certain product features to the forefront in the multimodal item representations, which benefits the outfit recommendation results. We demonstrated this on three datasets, improving over state-of-the-art results on an outfit compatibility prediction task and an outfit completion task.

As future work and to further improve the results, we would like to investigate neural architectures that still better recognise fine-grained fashion attributes in images, to benefit more from the attention-based fusion. Furthermore, we would like to design novel co-attention mechanisms which still better integrate fine-grained visual and textual attributes. 

\begin{acks}
  The first author is supported by a grant of the Research Foundation - Flanders (FWO) no. 1S55420N.
\end{acks}

\bibliographystyle{ACM-Reference-Format}
\bibliography{references}


\begin{thebibliography}{00}

%%% the Web address.

\ifx \showCODEN    \undefined \def \showCODEN     #1{\unskip}     \fi
\ifx \showDOI      \undefined \def \showDOI       #1{#1}\fi
\ifx \showISBNx    \undefined \def \showISBNx     #1{\unskip}     \fi
\ifx \showISBNxiii \undefined \def \showISBNxiii  #1{\unskip}     \fi
\ifx \showISSN     \undefined \def \showISSN      #1{\unskip}     \fi
\ifx \showLCCN     \undefined \def \showLCCN      #1{\unskip}     \fi
\ifx \shownote     \undefined \def \shownote      #1{#1}          \fi
\ifx \showarticletitle \undefined \def \showarticletitle #1{#1}   \fi
\ifx \showURL      \undefined \def \showURL       {\relax}        \fi
% invisible to TeX
\providecommand\bibfield[2]{#2}
\providecommand\bibinfo[2]{#2}
\providecommand\natexlab[1]{#1}
\providecommand\showeprint[2][]{arXiv:#2}

\bibitem[\protect\citeauthoryear{Han, Wu, Jiang, and Davis}{Han
  et~al\mbox{.}}{2017}]%
        {han2017}
\bibfield{author}{\bibinfo{person}{Xintong Han}, \bibinfo{person}{Zuxuan Wu},
  \bibinfo{person}{Yu-Gang Jiang}, {and} \bibinfo{person}{Larry~S Davis}.}
  \bibinfo{year}{2017}\natexlab{}.
\newblock \showarticletitle{Learning Fashion Compatibility with Bidirectional
  LSTMs}. In \bibinfo{booktitle}{{\em ACM Multimedia}}.
\newblock


\bibitem[\protect\citeauthoryear{He, Zhang, Ren, and Sun}{He
  et~al\mbox{.}}{2016b}]%
        {he2016}
\bibfield{author}{\bibinfo{person}{Kaiming He}, \bibinfo{person}{Xiangyu
  Zhang}, \bibinfo{person}{Shaoqing Ren}, {and} \bibinfo{person}{Jian Sun}.}
  \bibinfo{year}{2016}\natexlab{b}.
\newblock \showarticletitle{Deep Residual Learning for Image Recognition}.
\newblock \bibinfo{journal}{{\em 2016 IEEE Conference on Computer Vision and
  Pattern Recognition (CVPR)\/}} (\bibinfo{year}{2016}),
  \bibinfo{pages}{770--778}.
\newblock


\bibitem[\protect\citeauthoryear{He, Packer, and McAuley}{He
  et~al\mbox{.}}{2016a}]%
        {rhe2016}
\bibfield{author}{\bibinfo{person}{Ruining He}, \bibinfo{person}{Charles
  Packer}, {and} \bibinfo{person}{Julian McAuley}.}
  \bibinfo{year}{2016}\natexlab{a}.
\newblock \showarticletitle{Learning Compatibility Across Categories for
  Heterogeneous Item Recommendation}. In \bibinfo{booktitle}{{\em International
  Conference on Data Mining}}.
\newblock


\bibitem[\protect\citeauthoryear{Hsiao and Grauman}{Hsiao and Grauman}{2018}]%
        {hsiao2018}
\bibfield{author}{\bibinfo{person}{Wei{-}Lin Hsiao} {and}
  \bibinfo{person}{Kristen Grauman}.} \bibinfo{year}{2018}\natexlab{}.
\newblock \showarticletitle{Creating Capsule Wardrobes From Fashion Images}. In
  \bibinfo{booktitle}{{\em 2018 {IEEE} Conference on Computer Vision and
  Pattern Recognition, {CVPR} 2018, Salt Lake City, UT, USA, June 18-22,
  2018}}. \bibinfo{pages}{7161--7170}.
\newblock


\bibitem[\protect\citeauthoryear{Li, Cao, Zhu, and Luo}{Li
  et~al\mbox{.}}{2017}]%
        {li2017}
\bibfield{author}{\bibinfo{person}{Yuncheng Li}, \bibinfo{person}{Liangliang
  Cao}, \bibinfo{person}{Jiang Zhu}, {and} \bibinfo{person}{Jiebo Luo}.}
  \bibinfo{year}{2017}\natexlab{}.
\newblock \showarticletitle{Mining Fashion Outfit Composition Using an
  End-to-End Deep Learning Approach on Set Data}.
\newblock \bibinfo{journal}{{\em IEEE Transactions on Multimedia\/}}
  \bibinfo{volume}{19} (\bibinfo{year}{2017}), \bibinfo{pages}{1946--1955}.
\newblock


\bibitem[\protect\citeauthoryear{Lin, Ren, Chen, Ren, Ma, and de~Rijke}{Lin
  et~al\mbox{.}}{2019}]%
        {lin2019}
\bibfield{author}{\bibinfo{person}{Yujie Lin}, \bibinfo{person}{Pengjie Ren},
  \bibinfo{person}{Zhumin Chen}, \bibinfo{person}{Zhaochun Ren},
  \bibinfo{person}{Jun Ma}, {and} \bibinfo{person}{Maarten de Rijke}.}
  \bibinfo{year}{2019}\natexlab{}.
\newblock \showarticletitle{Improving Outfit Recommendation with Co-supervision
  of Fashion Generation}. In \bibinfo{booktitle}{{\em The World Wide Web
  Conference}} {\em (\bibinfo{series}{WWW '19})}. \bibinfo{publisher}{ACM},
  \bibinfo{pages}{1095--1105}.
\newblock


\bibitem[\protect\citeauthoryear{Lu, Yang, Batra, and Parikh}{Lu
  et~al\mbox{.}}{2016}]%
        {lu2016}
\bibfield{author}{\bibinfo{person}{Jiasen Lu}, \bibinfo{person}{Jianwei Yang},
  \bibinfo{person}{Dhruv Batra}, {and} \bibinfo{person}{Devi Parikh}.}
  \bibinfo{year}{2016}\natexlab{}.
\newblock \showarticletitle{Hierarchical Question-image Co-attention for Visual
  Question Answering}. In \bibinfo{booktitle}{{\em Proceedings of the 30th
  International Conference on Neural Information Processing Systems}} {\em
  (\bibinfo{series}{NIPS'16})}. \bibinfo{publisher}{Curran Associates Inc.},
  \bibinfo{pages}{289--297}.
\newblock


\bibitem[\protect\citeauthoryear{Nam, Ha, and Kim}{Nam et~al\mbox{.}}{2017}]%
        {nam2017}
\bibfield{author}{\bibinfo{person}{Hyeonseob Nam}, \bibinfo{person}{Jung-Woo
  Ha}, {and} \bibinfo{person}{Jeonghee Kim}.} \bibinfo{year}{2017}\natexlab{}.
\newblock \showarticletitle{Dual Attention Networks for Multimodal Reasoning
  and Matching}. In \bibinfo{booktitle}{{\em The IEEE Conference on Computer
  Vision and Pattern Recognition (CVPR)}}.
\newblock


\bibitem[\protect\citeauthoryear{Seo, Kembhavi, Farhadi, and Hajishirzi}{Seo
  et~al\mbox{.}}{2017}]%
        {seo2017}
\bibfield{author}{\bibinfo{person}{Min~Joon Seo}, \bibinfo{person}{Aniruddha
  Kembhavi}, \bibinfo{person}{Ali Farhadi}, {and} \bibinfo{person}{Hannaneh
  Hajishirzi}.} \bibinfo{year}{2017}\natexlab{}.
\newblock \showarticletitle{Bidirectional Attention Flow for Machine
  Comprehension}. In \bibinfo{booktitle}{{\em 5th International Conference on
  Learning Representations, {ICLR} 2017, Toulon, France, April 24-26, 2017,
  Conference Track Proceedings}}.
\newblock


\bibitem[\protect\citeauthoryear{Simo-Serra, Fidler, Moreno-Noguer, and
  Urtasun}{Simo-Serra et~al\mbox{.}}{2015}]%
        {simoserra2015}
\bibfield{author}{\bibinfo{person}{Edgar Simo-Serra}, \bibinfo{person}{Sanja
  Fidler}, \bibinfo{person}{Francesc Moreno-Noguer}, {and}
  \bibinfo{person}{Raquel Urtasun}.} \bibinfo{year}{2015}\natexlab{}.
\newblock \showarticletitle{Neuroaesthetics in fashion: Modeling the perception
  of fashionability.}. In \bibinfo{booktitle}{{\em CVPR}}.
  \bibinfo{publisher}{IEEE Computer Society}, \bibinfo{pages}{869--877}.
\newblock


\bibitem[\protect\citeauthoryear{Vasileva, Plummer, Dusad, Rajpal, Kumar, and
  Forsyth}{Vasileva et~al\mbox{.}}{2018}]%
        {vasileva2018}
\bibfield{author}{\bibinfo{person}{Mariya~I. Vasileva},
  \bibinfo{person}{Bryan~A. Plummer}, \bibinfo{person}{Krishna Dusad},
  \bibinfo{person}{Shreya Rajpal}, \bibinfo{person}{Ranjitha Kumar}, {and}
  \bibinfo{person}{David~A. Forsyth}.} \bibinfo{year}{2018}\natexlab{}.
\newblock \showarticletitle{Learning Type-Aware Embeddings for Fashion
  Compatibility}. In \bibinfo{booktitle}{{\em ECCV}}.
\newblock


\bibitem[\protect\citeauthoryear{Vaswani, Shazeer, Parmar, Uszkoreit, Jones,
  Gomez, Kaiser, and Polosukhin}{Vaswani et~al\mbox{.}}{2017}]%
        {vaswani2017}
\bibfield{author}{\bibinfo{person}{Ashish Vaswani}, \bibinfo{person}{Noam
  Shazeer}, \bibinfo{person}{Niki Parmar}, \bibinfo{person}{Jakob Uszkoreit},
  \bibinfo{person}{Llion Jones}, \bibinfo{person}{Aidan~N Gomez},
  \bibinfo{person}{Lukasz Kaiser}, {and} \bibinfo{person}{Illia Polosukhin}.}
  \bibinfo{year}{2017}\natexlab{}.
\newblock \showarticletitle{Attention is All you Need}.
\newblock In \bibinfo{booktitle}{{\em Advances in Neural Information Processing
  Systems 30}}, \bibfield{editor}{\bibinfo{person}{I.~Guyon},
  \bibinfo{person}{U.~V. Luxburg}, \bibinfo{person}{S.~Bengio},
  \bibinfo{person}{H.~Wallach}, \bibinfo{person}{R.~Fergus},
  \bibinfo{person}{S.~Vishwanathan}, {and} \bibinfo{person}{R.~Garnett}}
  (Eds.). \bibinfo{publisher}{Curran Associates, Inc.},
  \bibinfo{pages}{5998--6008}.
\newblock


\bibitem[\protect\citeauthoryear{Veit, Kovacs, Bell, McAuley, Bala, and
  Belongie}{Veit et~al\mbox{.}}{2015}]%
        {veit2015}
\bibfield{author}{\bibinfo{person}{Andreas Veit}, \bibinfo{person}{Balazs
  Kovacs}, \bibinfo{person}{Sean Bell}, \bibinfo{person}{Julian McAuley},
  \bibinfo{person}{Kavita Bala}, {and} \bibinfo{person}{Serge Belongie}.}
  \bibinfo{year}{2015}\natexlab{}.
\newblock \showarticletitle{Learning Visual Clothing Style with Heterogeneous
  Dyadic Co-Occurrences}. In \bibinfo{booktitle}{{\em Proceedings of the 2015
  IEEE International Conference on Computer Vision (ICCV)}} {\em
  (\bibinfo{series}{ICCV '15})}. \bibinfo{publisher}{IEEE Computer Society},
  \bibinfo{pages}{4642--4650}.
\newblock


\bibitem[\protect\citeauthoryear{Yang, He, Gao, Deng, and Smola}{Yang
  et~al\mbox{.}}{2016}]%
        {yang2016}
\bibfield{author}{\bibinfo{person}{Zichao Yang}, \bibinfo{person}{Xiaodong He},
  \bibinfo{person}{Jianfeng Gao}, \bibinfo{person}{Li Deng}, {and}
  \bibinfo{person}{Alexander~J. Smola}.} \bibinfo{year}{2016}\natexlab{}.
\newblock \showarticletitle{Stacked Attention Networks for Image Question
  Answering.}. In \bibinfo{booktitle}{{\em CVPR}}. \bibinfo{publisher}{IEEE
  Computer Society}, \bibinfo{pages}{21--29}.
\newblock


\bibitem[\protect\citeauthoryear{Yu, Yu, Fan, and Tao}{Yu
  et~al\mbox{.}}{2017}]%
        {yu2017}
\bibfield{author}{\bibinfo{person}{Zhou Yu}, \bibinfo{person}{Jun Yu},
  \bibinfo{person}{Jianping Fan}, {and} \bibinfo{person}{Dacheng Tao}.}
  \bibinfo{year}{2017}\natexlab{}.
\newblock \showarticletitle{Multi-modal Factorized Bilinear Pooling with
  Co-Attention Learning for Visual Question Answering}.
\newblock \bibinfo{journal}{{\em IEEE International Conference on Computer
  Vision (ICCV)\/}} (\bibinfo{year}{2017}), \bibinfo{pages}{1839--1848}.
\newblock


\end{thebibliography}

\appendix

\section{Dataset Item Types}

Table \ref{tab:types} gives an overview of the different item types in the Polyvore68K dataset and the types that remain in the Polyvore21K dataset after cleaning.

\begin{table}[h]
    \centering
    \begin{center}
    \resizebox{0.48\textwidth}{!}{%
    \begin{tabular}{| l | l |}
    \hline
    & \textbf{Item Types}\\
    \hline
    \multirow{3}{*}{\textbf{Polyvore68K}} & Accessories, All body, Bags, Bottoms, Hats, \\
    & Jewellery, Outerwear, Scarves, Shoes, Sun- \\
    & glasses, Tops\\
    \hline
    \multirow{8}{*}{\textbf{Polyvore21K}} & Accessories, Activewear, Baby, Bags and Wallets,\\
    & Belts, Boys, Cardigans and Vests, Clothing, Cos-\\
    & tumes, Cover-ups, Dresses, Eyewear, Girls, Gloves,\\
    & Hats, Hosiery and Socks, Jeans, Jewellery, Jumpsuits,\\
    & Juniors, Kids, Maternity, Outerwear, Pants, Scarves,\\
    & Shoes, Shorts, Skirts, Sleepwear, Suits, Sweaters and \\
    & Hoodies, Swimwear, Ties, Tops, Underwear, Watches,\\
    & Wedding Dresses \\
    \hline
    \end{tabular}}
    \end{center}
    \caption{Item types kept in the Polyvore68K and Polyvore21K datasets.}
    \label{tab:types}
\end{table}

\end{document}